\documentclass[10pt,twocolumn,letterpaper]{article}

\usepackage{cvpr}              

\usepackage[dvipsnames]{xcolor}

\usepackage{float}
\usepackage{caption}

\definecolor{cvprblue}{rgb}{0.21,0.49,0.74}
\usepackage[pagebackref,breaklinks,colorlinks,citecolor=cvprblue]{hyperref}

\def\paperID{*****} 
\def\confName{CVPR}
\def\confYear{2025}

\title{CellSymphony: Deciphering the molecular and phenotypic orchestration of cells with single-cell pathomics}

\author{%
\begin{tabular}{c}
Paul H. Acosta\textsuperscript{1,2,*} \quad Pingjun Chen\textsuperscript{1,2} \quad Simon P. Castillo\textsuperscript{1,2} \\
Maria Esther Salvatierra\textsuperscript{1} \quad Yinyin Yuan\textsuperscript{1,2,*} \quad Xiaoxi Pan\textsuperscript{1,2,*} \\
\small\textsuperscript{1}Translational Molecular Pathology Department, Division of Pathology and Laboratory Medicine, \\
\small The University of Texas MD Anderson Cancer Center, Houston, TX, United States \\
\small\textsuperscript{2}Institute for Data Science in Oncology, The University of Texas MD Anderson Cancer Center, Houston, TX, United States \\[4pt]
\small\textsuperscript{*}Corresponding authors: \texttt{phacosta@mdanderson.org, yyuan6@mdanderson.org, xpan7@mdanderson.org}
\end{tabular}
}

\begin{document}
\maketitle
\begin{abstract}

Xenium, a new spatial transcriptomics platform, enables subcellular-resolution profiling of complex tumor tissues. Despite the rich morphological information in histology images, extracting robust cell-level features and integrating them with spatial transcriptomics data remains a critical challenge. We introduce CellSymphony, a flexible multimodal framework that leverages foundation model-derived embeddings from both Xenium transcriptomic profiles and histology images at true single-cell resolution. By learning joint representations that fuse spatial gene expression with morphological context, CellSymphony achieves accurate cell type annotation and uncovers distinct microenvironmental niches across three cancer types. This work highlights the potential of foundation models and multimodal fusion for deciphering the physiological and phenotypic orchestration of cells within complex tissue ecosystems.

\end{abstract}

\begin{figure*}[htbp]
  \centering

  \begin{subfigure}{0.48\textwidth}
    \includegraphics[width=\linewidth, trim=0 60 0 0, clip]{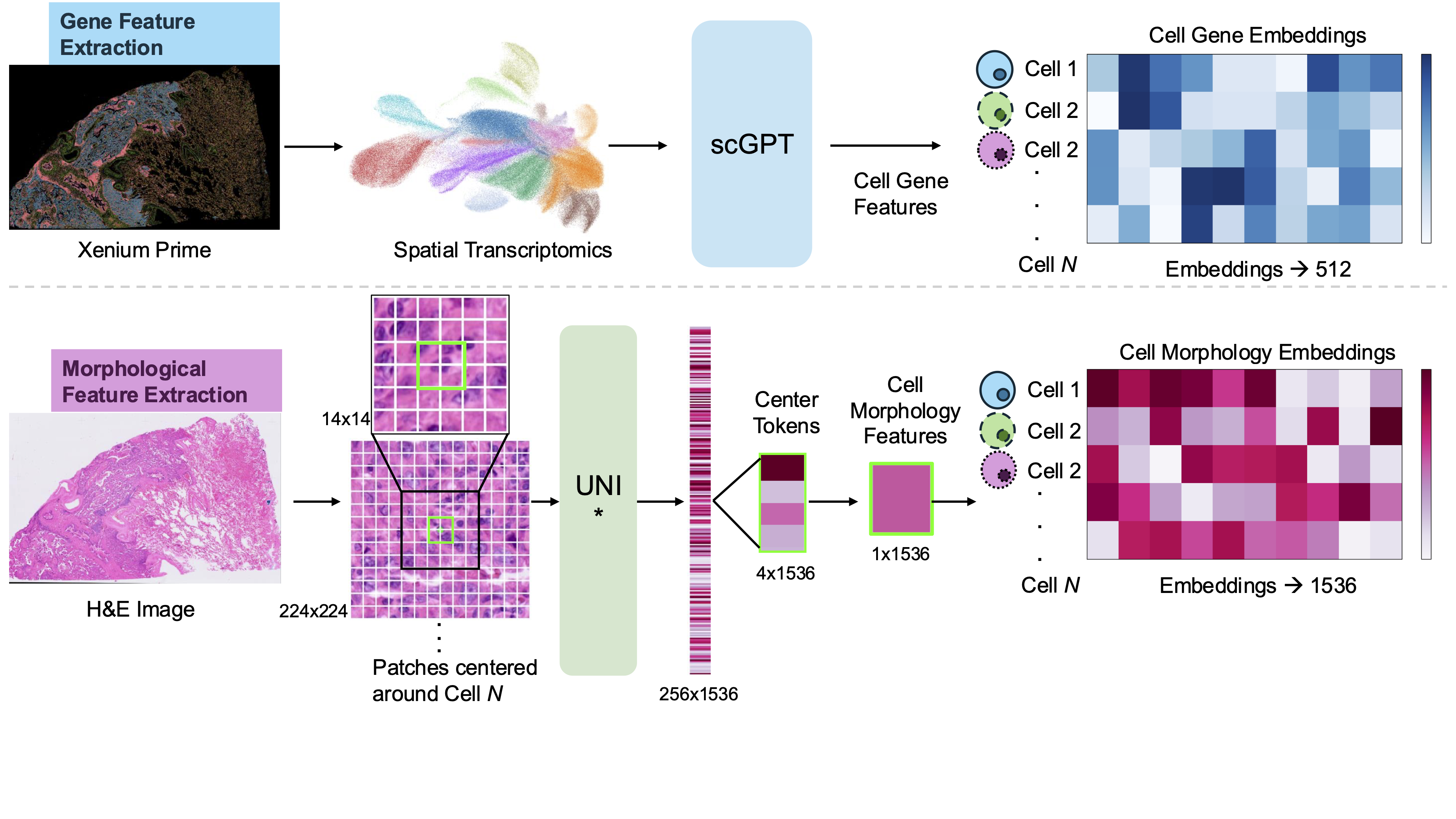}
    \put(-250,135){\textbf{(a)}} 
    \phantomsubcaption
    \label{fig:fig1a}
  \end{subfigure}
  \hfill
  \begin{subfigure}{0.48\textwidth}
    \includegraphics[width=\linewidth, trim=0 60 0 0, clip]{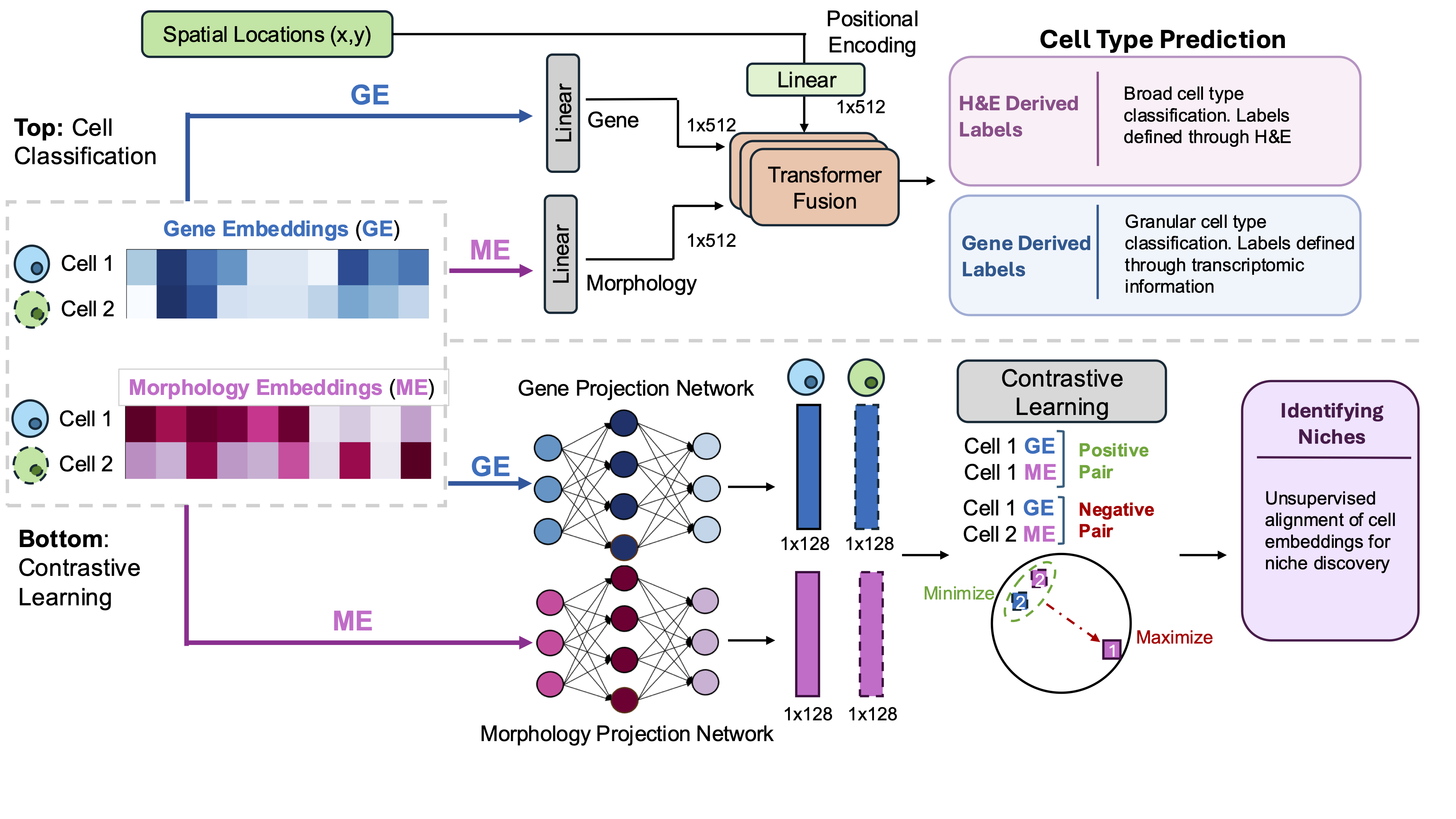}
    \put(-250,135){\textbf{(b)}}
    \phantomsubcaption
    \label{fig:fig1b}
  \end{subfigure}

  \caption{Overview of the CellSymphony framework. 
  \textbf{(a)} Cell embeddings are derived from Xenium transcriptomes using scGPT (top) and from histology images via a modified UNI* model (bottom). 
  \textbf{(b)} Architectures of two deep learning approaches: transformer-based cell type annotation (top) and contrastive learning for cross-modal alignment of transcriptomic and morphological features (bottom).}
  \label{fig:overview}
\end{figure*}

\section{Introduction}
\label{sec:intro}
The tumor microenvironment is a complex ecosystem where diverse cells spatially and molecularly interact, shaping cancer progression \cite{anderson2020tumor, de2023evolving}. Recent spatial transcriptomic advances enable high-resolution gene expression profiling, revealing tumor spatial organization and molecular heterogeneity \cite{moses2022museum, karimi2024method}. Among these technologies, the Xenium platform allows subcellular transcriptomic profiling of up to 5,000 genes, opening new avenues to investigate cell phenotypes, their communication, and the emerging tissue architecture \cite{janesick2023high}.

Although histology images provide rich contextual and morphological information about cells and tissue structures, they lack molecular resolution. Conversely, Xenium offers molecular insights, but often struggles with accurate cell classification in heterogeneous environments, particularly when cell boundaries are ambiguous or when expression profiles are sparse \cite{ozirmak2024comparison}. Hence, harnessing these two modalities results paramount to achieve a comprehensive understanding of cellular phenotypes in situ. Current methods often rely on indirect inference of genomic alterations \cite{ru2023estimation} or marker gene-based \cite{jiang2024meti} identification, which could be unreliable in complex tumor microenvironments. This highlights the necessity of integrating orthogonal modalities, histology and spatial transcriptomics, to improve tumor cell annotation. Several recent studies have sought to integrate histology images and spatial transcriptomics data, particularly from the Visium platform, to enhance downstream analyses such as gene imputation from morphology \cite{he2020integrating,xie2023spatially,zhang2024inferring,wang2025m2ost}, cell phenotyping \cite{zhao2024hist2cell, jiang2024meti}, and tissue characterization \cite{bao2022integrative, hu2023deciphering, he2025starfysh}. However, these efforts were mainly focused on spot–patch or spot–cell levels and did not fully exploit the rich, cell-level resolution. Consequently, the cell-level integration of histology and transcriptomics remains largely underexplored. 

Foundation models have advanced both morphological \cite{chen2024towards, xu2024whole} and molecular analyses \cite{theodoris2023transfer,cui2024scgpt}. Trained on extensive data across cancer types, these models learn generalizable representations that sharpen downstream tasks. Built on pathology and single-cell foundation models, UNI2 \cite{chen2024towards} and scGPT \cite{cui2024scgpt}, we introduce \textbf{CellSymphony} (Fig.~\ref{fig:overview}), a multimodal framework that integrates morphological and transcriptomic information at single-cell resolution to enhance spatial tissue characterization and derive biological insights into the highly orchestrated tumor ecosystem. 
Our experimental results demonstrate that integrating gene and morphological embeddings improves cell type annotation accuracy and enhances the discovery of spatial tissue subtypes and microenvironmental niches.

\section{Methods}
\label{sec:method}

Designed for Xenium Prime 5k spatial transcriptomics data paired with H\&E images, the CellSymphony framework first extracts deep feature embeddings from each modality using foundation models. These embeddings then drive two core capabilities: (1) multimodal transformer-based cell type classification, and (2) contrastive learning to align morphology and transcriptomics.

\subsection{Cell embedding at morphological and transcriptomic level}
\label{subsec:morphology_embeddings}
We derived a paired and modality-specific embedding for each cell (Fig.~\ref{fig:fig1a}). At the transcriptomic level, the cell’s gene counts vector is fed into the pre-trained scGPT model, producing a fixed transcriptomic embedding (Fig.~\ref{fig:fig1a}, \textit{Top}). At the morphological level, a 224$\times$224 H\&E patch (0.5µm/px) centered on the same cell's nucleus is passed through UNI2, generating the morphological embedding. Rather than using the model’s default tile feature representation, we extracted intermediate spatial tokens from UNI2 and aggregate those nearest the nucleus to generate a 1536-dimensional cell-specific morphology embedding, allowing flexible spatial context (Fig.~\ref{fig:fig1a}, \textit{Bottom}). Together, these matched morphology and RNA vectors form the joint input for our cross-modal analyses.

\subsection{Transformer-based cell type classification}
We trained transformer-based models to classify cell types from gene expression and morphology embeddings derived from Xenium Prime data (Fig.~\ref{fig:fig1b}, \textit{Top}). Cell annotations were sourced from either (1) SingleR \cite{aran2019reference}, which uses reference-based transcriptomic classification, or (2) AI-sTIL \cite{abduljabbar2020geospatial, pan2025tmeseg}, an H\&E-based segmentation and classification pipeline. Models were trained separately per label source, excluding ambiguous or low-confidence annotations, and data was split into 80\% training and 20\% testing.

We evaluated four transformer architectures: (1) a \textbf{Unimodal Transformer} trained on gene embeddings (GEB) alone; (2) a \textbf{Spatial Transformer} that adds a second input token for spatial coordinates using sinusoidal encoding; (3) a \textbf{Dual-Modality Transformer}, which processes gene and morphology embeddings (MEB) as separate input tokens, each tagged with a modality-specific token; and (4) a \textbf{Multi-Input Transformer}, which incorporates all three inputs (gene, morphology, spatial) using relative positional attention to guide interactions between tokens. Each embedding was projected into a shared dimensional space, then passed through a 6-layer transformer fusion encoder. The outputs were mean-pooled and classified using a linear head.
All models were trained for 20 epochs using AdamW (learning rate $3 \times 10^{-5}$, batch size 64, weight decay 0.01), with cross-entropy loss and class weights to address label imbalance. 

\subsection{Contrastive learning representation between morphology and transcriptomics}
To align the transcriptomic and morphological embeddings in a shared space, we implemented an unsupervised contrastive learning framework (Fig.~\ref{fig:fig1b}, \textit{Bottom}). As described previously, cell-specific features were extracted from H\&E and Xenium data using UNI2 and scGPT, respectively. Each modality was passed through a separate projection network to map into a 128-dimensional latent space.

The networks were trained using InfoNCE loss \cite{oord2018representation}, which pulls matched embeddings (same cell) together while pushing unmatched pairs apart. Loss was computed bidirectionally, from morphology to transcriptomics and vice versa, on each batch of cells. Downstream analysis focused on the individual projections after training.

We evaluated the resulting features using UMAP \cite{McInnes2018}, clustering, and spatial visualization to assess improvements in the detection of biological signal and niches, particularly in morphology-derived representations.

\section{Experiments and Results}
\label{sec:result}
\subsection{Dataset}
We used lung ($n_\mathrm{cells}=244{,}659$), breast ($n_\mathrm{cells}=461{,}094$), and prostate ($n_\mathrm{cells}=151{,}665$) \textit{Xenium Prime 5k} datasets from 10x Genomics. Each sample includes gene expression from the Human Pan Tissue \& Pathways panel, spatial cell coordinates, and matched high-resolution H\&E images.

\begin{center}
\scriptsize
\resizebox{\columnwidth}{!}{%
\begin{tabular}{lccccccccc}
\toprule
Tissue & $P_{\text{fib}}$ & $R_{\text{fib}}$ & $F1_{\text{fib}}$ & $P_{\text{lym}}$ & $R_{\text{lym}}$ & $F1_{\text{lym}}$ & $P_{\text{tum}}$ & $R_{\text{tum}}$ & $F1_{\text{tum}}$ \\
\midrule
\addlinespace
\multicolumn{10}{l}{\textit{Unimodal Transformer (GEB Trained)}} \\
\midrule
Breast & 0.71 & 0.65 & 0.68 & 0.55 & 0.65 & 0.60 & 0.85 & 0.85 & 0.85 \\
Lung & 0.80 & 0.79 & 0.79 & 0.56 & 0.65 & 0.60 & 0.83 & 0.80 & 0.82 \\
Prostate & 0.88 & 0.83 & 0.86 & 0.28 & 0.09 & 0.14 & 0.78 & 0.85 & 0.82 \\
\addlinespace
\multicolumn{10}{l}{\textit{Spatial Transformer (GEB Trained)}} \\
\midrule
Breast & 0.70 & 0.70 & 0.70 & 0.58 & 0.59 & 0.59 & 0.85 & 0.84 & 0.85 \\
Lung & 0.79 & 0.82 & 0.81 & 0.56 & 0.63 & 0.59 & 0.87 & 0.79 & 0.82 \\
Prostate & 0.88 & 0.85 & 0.86 & 0.31 & 0.09 & 0.14 & 0.80 & 0.85 & 0.82 \\
\addlinespace
\multicolumn{10}{l}{\textit{Dual-Modality Transformer (GEB + MEB Trained)}} \\
\midrule
Breast & 0.90 & 0.90 & \textbf{0.90} & 0.90 & 0.87 & \textbf{0.88} & 0.92 & 0.95 & \textbf{0.93} \\
Lung & 0.90 & 0.93 & \textbf{0.91} & 0.94 & 0.81 & \textbf{0.87} & 0.91 & 0.91 & \textbf{0.91} \\
Prostate & 0.93 & 0.89 & \textbf{0.91} & 0.71 & 0.66 & \textbf{0.68} & 0.86 & 0.92 & \textbf{0.89} \\
\addlinespace
\multicolumn{10}{l}{\textit{Multi-Input Transformer (GEB + MEB Trained)}} \\
\midrule
Breast & 0.93 & 0.85 & 0.89 & 0.84 & 0.93 & 0.88 & 0.92 & 0.95 & \textbf{0.93} \\
Lung & 0.93 & 0.87 & 0.90 & 0.77 & 0.96 & 0.85 & 0.90 & 0.92 & \textbf{0.91} \\
Prostate & 0.93 & 0.87 & 0.90 & 0.30 & 0.85 & 0.44 & 0.85 & 0.91 & 0.88 \\
\bottomrule
\end{tabular}%
}
\captionof{table}{Performance metrics for AI-sTIL predictions across tissues. \textbf{Abbreviations:} P = Precision, R = Recall, F1 = F1-score; fib = Fibroblast, lym = Lymphocyte, tum = Tumor.}
\label{tab:AI-sTIL_results}
\end{center}

\begin{table*}[t]
\centering
\scriptsize
\resizebox{\textwidth}{!}{%
\begin{tabular}{lcccccccccccccccccc}
\toprule
Tissue & $P_{b\_c}$ & $R_{b\_c}$ & $F1_{b\_c}$ & $P_{end}$ & $R_{end}$ & $F1_{end}$ & $P_{epi}$ & $R_{epi}$ & $F1_{epi}$ & $P_{fib}$ & $R_{fib}$ & $F1_{fib}$ & $P_{mac}$ & $R_{mac}$ & $F1_{mac}$ & $P_{t\_c}$ & $R_{t\_c}$ & $F1_{t\_c}$ \\
\midrule
\addlinespace
\multicolumn{19}{l}{\textit{Unimodal Transformer (GEB Trained)}} \\
\midrule
Breast & 0.77 & 0.79 & 0.78 & 0.80 & 0.76 & 0.78 & 0.94 & 0.97 & 0.95 & 0.85 & 0.80 & 0.82 & 0.82 & 0.83 & 0.82 & 0.89 & 0.87 & 0.88 \\
Lung & 0.94 & 0.87 & 0.90 & 0.94 & 0.93 & 0.94 & 0.97 & 0.97 & 0.97 & 0.93 & 0.92 & 0.92 & 0.88 & 0.95 & 0.92 & 0.93 & 0.94 & 0.93 \\
Prostate & 0.54 & 0.55 & 0.55 & 0.84 & 0.76 & 0.80 & 0.92 & 0.97 & 0.94 & 0.89 & 0.85 & 0.87 & 0.89 & 0.80 & 0.84 & 0.84 & 0.77 & 0.81 \\
\addlinespace
\multicolumn{19}{l}{\textit{Spatial Transformer (GEB Trained)}} \\
\midrule
Breast & 0.81 & 0.75 & 0.78 & 0.81 & 0.76 & 0.78 & 0.94 & 0.97 & 0.95 & 0.81 & 0.82 & 0.82 & 0.82 & 0.82 & 0.82 & 0.88 & 0.89 & 0.88 \\
Lung & 0.88 & 0.92 & 0.90 & 0.94 & 0.93 & 0.93 & 0.97 & 0.98 & 0.97 & 0.92 & 0.93 & \textbf{0.93} & 0.93 & 0.93 & 0.93 & 0.94 & 0.92 & 0.93 \\
Prostate & 0.76 & 0.49 & 0.59 & 0.88 & 0.75 & 0.81 & 0.93 & 0.96 & 0.95 & 0.88 & 0.87 & 0.88 & 0.80 & 0.85 & 0.82 & 0.84 & 0.79 & 0.81 \\
\addlinespace
\multicolumn{19}{l}{\textit{Dual-Modality Transformer (GEB + MEB Trained)}} \\
\midrule
Breast & 0.78 & 0.83 & \textbf{0.81} & 0.94 & 0.68 & \textbf{0.79} & 0.94 & 0.98 & \textbf{0.96} & 0.84 & 0.82 & \textbf{0.83} & 0.85 & 0.82 & \textbf{0.83} & 0.90 & 0.89 & \textbf{0.89} \\
Lung & 0.91 & 0.92 & 0.91 & 0.95 & 0.92 & 0.94 & 0.97 & 0.98 & 0.97 & 0.94 & 0.91 & \textbf{0.93} & 0.91 & 0.94 & 0.93 & 0.93 & 0.95 & 0.94 \\
Prostate & 0.78 & 0.49 & \textbf{0.60} & 0.94 & 0.74 & \textbf{0.83} & 0.92 & 0.98 & \textbf{0.95} & 0.91 & 0.88 & \textbf{0.89} & 0.90 & 0.82 & \textbf{0.86} & 0.88 & 0.76 & \textbf{0.82} \\
\addlinespace
\multicolumn{19}{l}{\textit{Multi-Input Transformer (GEB + MEB Trained)}} \\
\midrule
Breast & 0.78 & 0.82 & 0.80 & 0.69 & 0.82 & 0.75 & 0.97 & 0.95 & \textbf{0.96} & 0.79 & 0.84 & 0.82 & 0.79 & 0.86 & 0.82 & 0.93 & 0.84 & 0.88 \\
Lung & 0.95 & 0.97 & \textbf{0.96} & 0.98 & 0.98 & \textbf{0.98} & 1.00 & 0.99 & \textbf{0.99} & 0.92 & 0.95 & \textbf{0.93} & 0.96 & 0.98 & \textbf{0.97} & 0.98 & 0.97 & \textbf{0.97} \\
Prostate & 0.21 & 0.74 & 0.32 & 0.82 & 0.77 & 0.79 & 0.95 & 0.93 & 0.94 & 0.94 & 0.83 & 0.88 & 0.74 & 0.85 & 0.79 & 0.81 & 0.76 & 0.78 \\
\bottomrule
\end{tabular}
}
\caption{Performance metrics for SingleR predictions across tissues. \textbf{Abbreviations:} P = Precision, R = Recall, F1 = F1-score; subscripts: b\_c = B cell, end = Endothelial, epi = Epithelial, fib = Fibroblast, mac = Macrophage, t\_c = T cell. GEB: gene embeddings and MEB: morphological embeddings.}
\label{tab:singler_results}
\end{table*}

\subsection{Performance of cell type classification}

We evaluated classification models in three distinct tissue types: breast cancer, with large clusters of invasive and in situ carcinoma interspersed with normal ducts; lung cancer, where malignant epithelial cells are intermixed with stroma, immune cells, and tubular structures (e.g., bronchi); and prostate cancer, characterized by low immune infiltration and morphologically similar normal and cancerous glands.

Against morphology-based ground truth (AI-sTIL labels; Table~\ref{tab:AI-sTIL_results}), spatial encoding provided only a marginal improvement over the Unimodal Transformer. In contrast, the Dual-Modality Transformer, which integrated both gene expression and morphology, yielded the strongest performance. Adding spatial input via the (Multi-Input Transformer) did not deliver further gains, suggesting that once robust gene and morphology embeddings are in place, spatial information offers limited additional benefit to single cell annotation.

Using transcriptomic ground truth (SingleR; Table~\ref{tab:singler_results}), the Dual-Modality Transformer consistently improved classification across cell types, achieving the highest F1 scores for B cells in lung (0.91), T cells in lung (0.94), and fibroblasts in prostate (0.86). Incorporating spatial context (Multi-Input Transformer) further improved performance in lung—particularly for epithelial cells (F1 = 0.99), B cells (F1 = 0.96), T cells (F1 = 0.97), and macrophages (F1 = 0.97), highlighting the importance of spatial organization in this tissue. In contrast, spatial input had a limited impact in breast and prostate, where performance gains were modest or inconsistent. Notably, B cell classification in prostate dropped markedly (n = 313), likely reflecting class imbalance or sensitivity to spatial noise.

\begin{figure*}[htbp]
  \centering

  \begin{subfigure}{\textwidth}
    \includegraphics[width=\linewidth, trim=0 1210 0 0, clip]{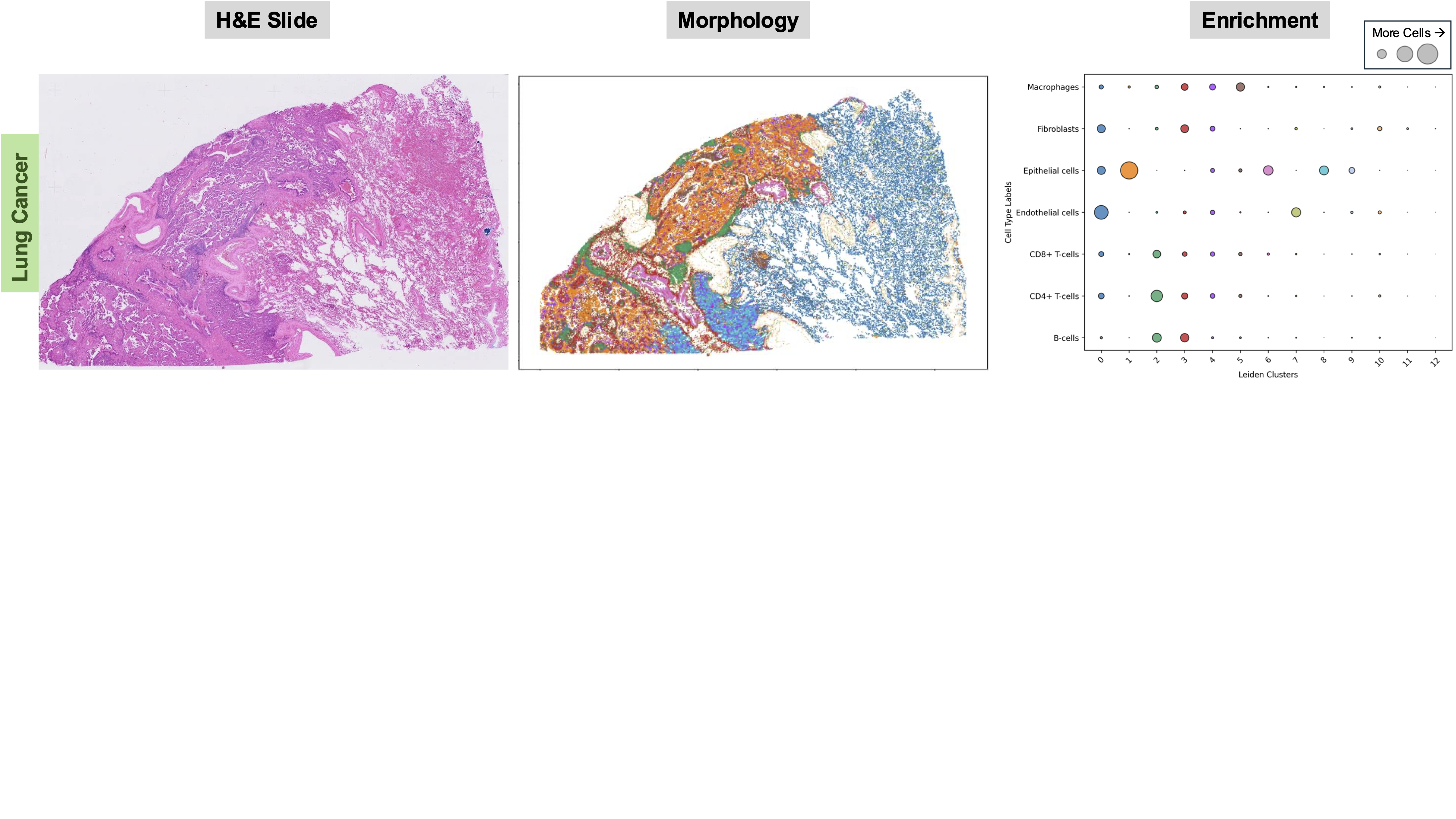}
    \put(-495,105){\textbf{(a)}}
    \phantomsubcaption
    \label{fig:fig2a}
  \end{subfigure}

  \vspace{0.1em}

  \begin{subfigure}{\textwidth}
    \includegraphics[width=\linewidth,trim=0 1390 0 0, clip]{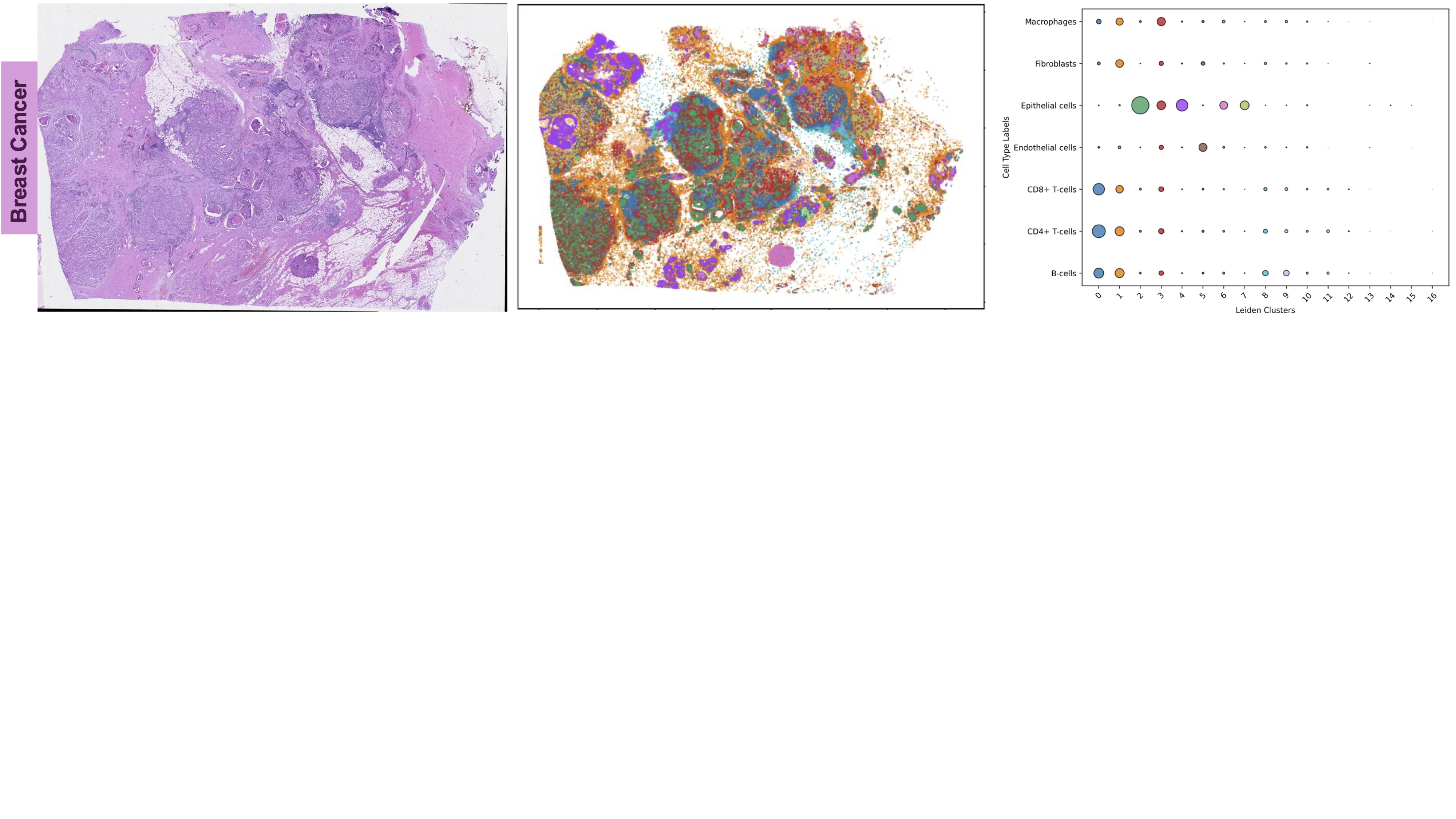}
    \put(-495,105){\textbf{(b)}}
    \phantomsubcaption
    \label{fig:fig2b}
  \end{subfigure}

  \vspace{0.1em}

  \begin{subfigure}{\textwidth}
    \includegraphics[width=\linewidth,trim=0 1400 0 0, clip]{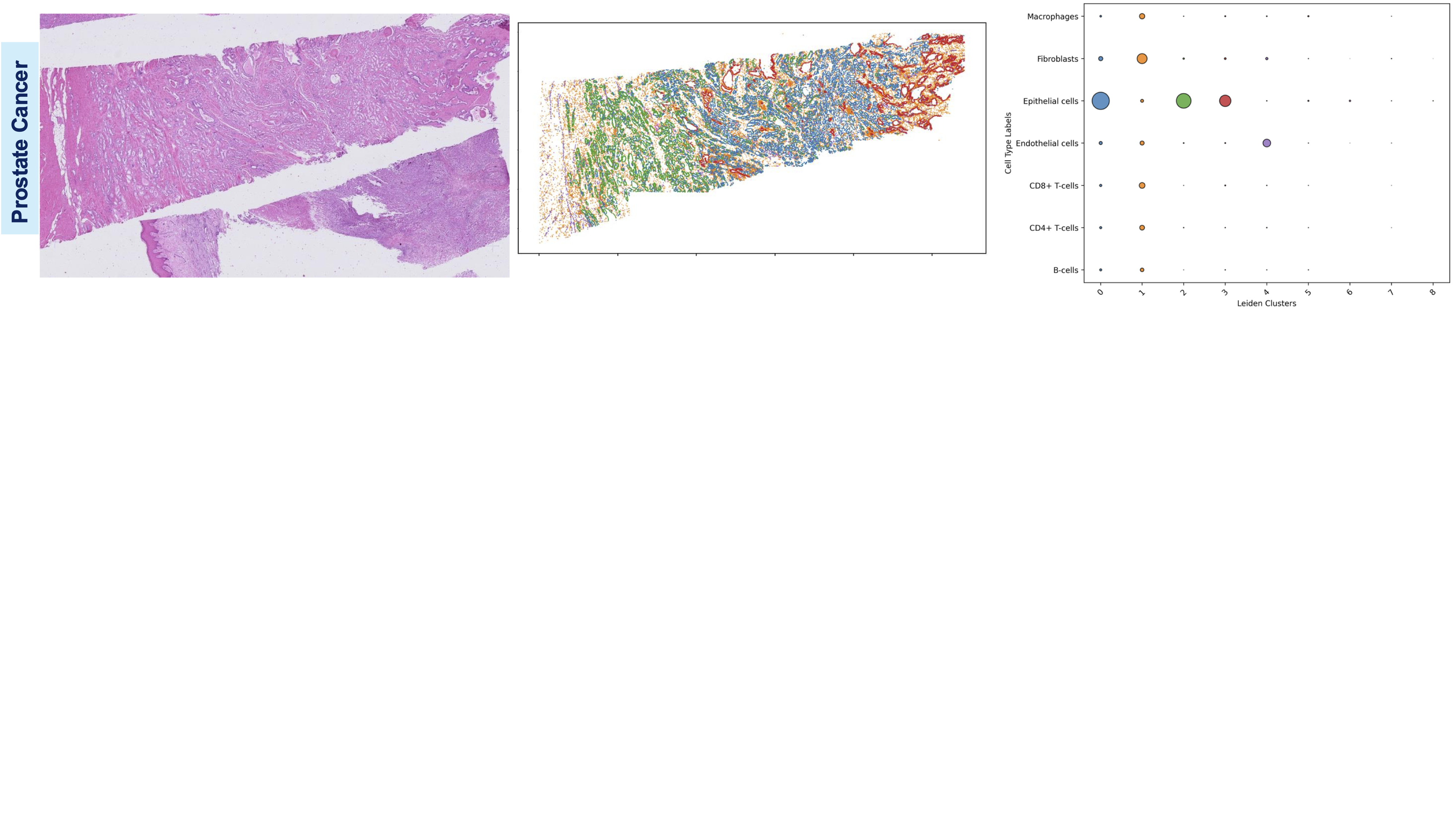}
    \put(-495,105){\textbf{(c)}}
    \phantomsubcaption
    \label{fig:fig2c}
  \end{subfigure}

  \caption{Morphology projection clustering after contrastive training for lung~\textbf{(a)}, breast~\textbf{(b)} and prostate~\textbf{(c)} samples. Each panel shows (from left to right):H\&E image, the morphological patterns defined by each cluster, and the enrichment of cell classes within those clusters.}
  \label{fig:contrastive_clustering}
\end{figure*}

\subsection{Enhancement of microenvironmental niche discovery through contrastive learning}

While supervised classification achieved high accuracy, its reliance on predefined labels limits the discovery of nuanced biological structure. To overcome this, we applied a self-supervised contrastive learning framework to align gene expression and morphology into a shared space and evaluated the resulting morphology-derived projections.

Across tissues, contrast-trained morphology projections revealed refined spatial and phenotypic structure. In lung cancer, lymphocyte-rich niches became sharply defined, particularly in cluster 2, with improved spatial coherence and cleaner enrichment for B and T cells (Fig.~\ref{fig:fig2a}). In breast cancer, projections better delineated macrophage and fibroblast compartments and further stratified epithelial subtypes in multiple clusters (Fig.~\ref{fig:fig2b}). In prostate cancer, projections captured differentiation gradients among epithelial regions and provided clearer separation of glandular structures, aligning with known histological variation (Fig.~\ref{fig:fig2c}).

Together, these findings demonstrate that contrastive learning improves the ability of morphology-derived features to resolve complex tissue architecture, enabling unsupervised identification of immune-rich regions and tumor heterogeneity without requiring gene expression at inference.

\section{Discussion and Conclusion}
\label{sec:discon}

In this study, we introduced \textbf{CellSymphony}, a multimodal framework for integrating spatial transcriptomics and histological morphology at the single-cell level. Applied to Xenium Prime 5k data from lung, breast, and prostate cancer tissues, CellSymphony leverages multimodal fusion based on foundation models (scGPT and UNI2), empowering both supervised classification and contrastive learning.

A central strength of CellSymphony is its ability to operate at true single-cell resolution while flexibly fusing and aligning morphology and gene expression. The framework enables high classification performance across annotation sources (molecular-based SingleR and morphology-based AI-sTIL), and contrastive learning improves the biological structure captured by morphology alone—particularly in lymphocyte-rich and tumor epithelial regions.

CellSymphony provides a flexible foundation for future multimodal spatial analyses. Its modular design enables incorporation of additional data types such as protein imaging or multiplexed RNA assays, and it can be adapted to incorporate spatial graphs or biological priors like gene ontology and regulatory networks. By aligning diverse modalities at single-cell resolution, CellSymphony opens new directions for investigating the tumor microenvironment and spatial cellular organization.

{
    \small
    \bibliographystyle{ieeenat_fullname}
    \bibliography{main}

\begin{thebibliography}{25}
\providecommand{\natexlab}[1]{#1}
\providecommand{\url}[1]{\texttt{#1}}
\expandafter\ifx\csname urlstyle\endcsname\relax
  \providecommand{\doi}[1]{doi: #1}\else
  \providecommand{\doi}{doi: \begingroup \urlstyle{rm}\Url}\fi

\bibitem[AbdulJabbar et~al.(2020)AbdulJabbar, Raza, Rosenthal, Jamal-Hanjani, Veeriah, Akarca, Lund, Moore, Salgado, Al~Bakir, et~al.]{abduljabbar2020geospatial}
Khalid AbdulJabbar, Shan E~Ahmed Raza, Rachel Rosenthal, Mariam Jamal-Hanjani, Selvaraju Veeriah, Ayse Akarca, Tom Lund, David~A Moore, Roberto Salgado, Maise Al~Bakir, et~al.
\newblock Geospatial immune variability illuminates differential evolution of lung adenocarcinoma.
\newblock \emph{Nature medicine}, 26\penalty0 (7):\penalty0 1054--1062, 2020.

\bibitem[Anderson and Simon(2020)]{anderson2020tumor}
Nicole~M Anderson and M~Celeste Simon.
\newblock The tumor microenvironment.
\newblock \emph{Current biology}, 30\penalty0 (16):\penalty0 R921--R925, 2020.

\bibitem[Aran et~al.(2019)Aran, Looney, Liu, Wu, Fong, Hsu, Chak, Naikawadi, Wolters, Abate, et~al.]{aran2019reference}
Dvir Aran, Agnieszka~P Looney, Leqian Liu, Esther Wu, Valerie Fong, Austin Hsu, Suzanna Chak, Ram~P Naikawadi, Paul~J Wolters, Adam~R Abate, et~al.
\newblock Reference-based analysis of lung single-cell sequencing reveals a transitional profibrotic macrophage.
\newblock \emph{Nature immunology}, 20\penalty0 (2):\penalty0 163--172, 2019.

\bibitem[Bao et~al.(2022)Bao, Deng, Wan, Shen, Wang, Dai, Altschuler, and Wu]{bao2022integrative}
Feng Bao, Yue Deng, Sen Wan, Susan~Q Shen, Bo Wang, Qionghai Dai, Steven~J Altschuler, and Lani~F Wu.
\newblock Integrative spatial analysis of cell morphologies and transcriptional states with muse.
\newblock \emph{Nature biotechnology}, 40\penalty0 (8):\penalty0 1200--1209, 2022.

\bibitem[Chen et~al.(2024)Chen, Ding, Lu, Williamson, Jaume, Song, Chen, Zhang, Shao, Shaban, et~al.]{chen2024towards}
Richard~J Chen, Tong Ding, Ming~Y Lu, Drew~FK Williamson, Guillaume Jaume, Andrew~H Song, Bowen Chen, Andrew Zhang, Daniel Shao, Muhammad Shaban, et~al.
\newblock Towards a general-purpose foundation model for computational pathology.
\newblock \emph{Nature Medicine}, 30\penalty0 (3):\penalty0 850--862, 2024.

\bibitem[Cui et~al.(2024)Cui, Wang, Maan, Pang, Luo, Duan, and Wang]{cui2024scgpt}
Haotian Cui, Chloe Wang, Hassaan Maan, Kuan Pang, Fengning Luo, Nan Duan, and Bo Wang.
\newblock scgpt: toward building a foundation model for single-cell multi-omics using generative ai.
\newblock \emph{Nature Methods}, 21\penalty0 (8):\penalty0 1470--1480, 2024.

\bibitem[De~Visser and Joyce(2023)]{de2023evolving}
Karin~E De~Visser and Johanna~A Joyce.
\newblock The evolving tumor microenvironment: From cancer initiation to metastatic outgrowth.
\newblock \emph{Cancer cell}, 41\penalty0 (3):\penalty0 374--403, 2023.

\bibitem[He et~al.(2020)He, Bergenstr{\aa}hle, Stenbeck, Abid, Andersson, Borg, Maaskola, Lundeberg, and Zou]{he2020integrating}
Bryan He, Ludvig Bergenstr{\aa}hle, Linnea Stenbeck, Abubakar Abid, Alma Andersson, {\AA}ke Borg, Jonas Maaskola, Joakim Lundeberg, and James Zou.
\newblock Integrating spatial gene expression and breast tumour morphology via deep learning.
\newblock \emph{Nature biomedical engineering}, 4\penalty0 (8):\penalty0 827--834, 2020.

\bibitem[He et~al.(2025)He, Jin, Nazaret, Shi, Chen, Rampersaud, Dhillon, Valdez, Friend, Fan, et~al.]{he2025starfysh}
Siyu He, Yinuo Jin, Achille Nazaret, Lingting Shi, Xueer Chen, Sham Rampersaud, Bahawar~S Dhillon, Izabella Valdez, Lauren~E Friend, Joy~Linyue Fan, et~al.
\newblock Starfysh integrates spatial transcriptomic and histologic data to reveal heterogeneous tumor--immune hubs.
\newblock \emph{Nature Biotechnology}, 43\penalty0 (2):\penalty0 223--235, 2025.

\bibitem[Hu et~al.(2023)Hu, Coleman, Zhang, Lee, Kadara, Wang, and Li]{hu2023deciphering}
Jian Hu, Kyle Coleman, Daiwei Zhang, Edward~B Lee, Humam Kadara, Linghua Wang, and Mingyao Li.
\newblock Deciphering tumor ecosystems at super resolution from spatial transcriptomics with tesla.
\newblock \emph{Cell systems}, 14\penalty0 (5):\penalty0 404--417, 2023.

\bibitem[Janesick et~al.(2023)Janesick, Shelansky, Gottscho, Wagner, Williams, Rouault, Beliakoff, Morrison, Oliveira, Sicherman, et~al.]{janesick2023high}
Amanda Janesick, Robert Shelansky, Andrew~D Gottscho, Florian Wagner, Stephen~R Williams, Morgane Rouault, Ghezal Beliakoff, Carolyn~A Morrison, Michelli~F Oliveira, Jordan~T Sicherman, et~al.
\newblock High resolution mapping of the tumor microenvironment using integrated single-cell, spatial and in situ analysis.
\newblock \emph{Nature communications}, 14\penalty0 (1):\penalty0 8353, 2023.

\bibitem[Jiang et~al.(2024)Jiang, Liu, Qin, Chen, Wu, Pizzi, Lazcano, Yamashita, Xu, Pei, et~al.]{jiang2024meti}
Jiahui Jiang, Yunhe Liu, Jiangjiang Qin, Jianfeng Chen, Jingjing Wu, Melissa~P Pizzi, Rossana Lazcano, Kohei Yamashita, Zhiyuan Xu, Guangsheng Pei, et~al.
\newblock Meti: deep profiling of tumor ecosystems by integrating cell morphology and spatial transcriptomics.
\newblock \emph{Nature communications}, 15\penalty0 (1):\penalty0 7312, 2024.

\bibitem[Karimi et~al.(2024)Karimi, Simo, Milet, TE, ALSH, QU, AIL, ABS, ALIND, GOODMA, et~al.]{karimi2024method}
ELHAM Karimi, N Simo, N Milet, W TE, A ALSH, ND QU, L AIL, R ABS, A ALIND, ND~MORRIS GOODMA, et~al.
\newblock Method of the year 2024: spatial proteomics.
\newblock \emph{Nat Methods}, 21:\penalty0 2195--2196, 2024.

\bibitem[McInnes et~al.(2018)McInnes, Healy, Saul, and Großberger]{McInnes2018}
Leland McInnes, John Healy, Nathaniel Saul, and Lukas Großberger.
\newblock Umap: Uniform manifold approximation and projection.
\newblock \emph{Journal of Open Source Software}, 3\penalty0 (29):\penalty0 861, 2018.

\bibitem[Moses and Pachter(2022)]{moses2022museum}
Lambda Moses and Lior Pachter.
\newblock Museum of spatial transcriptomics.
\newblock \emph{Nature methods}, 19\penalty0 (5):\penalty0 534--546, 2022.

\bibitem[Oord et~al.(2018)Oord, Li, and Vinyals]{oord2018representation}
Aaron van~den Oord, Yazhe Li, and Oriol Vinyals.
\newblock Representation learning with contrastive predictive coding.
\newblock \emph{arXiv preprint arXiv:1807.03748}, 2018.

\bibitem[Ozirmak~Lermi et~al.(2024)Ozirmak~Lermi, Molina~Ayala, Hernandez, Lu, Khan, Serrano, Lubo, Hamana, Tomczak, Barnes, et~al.]{ozirmak2024comparison}
Nejla Ozirmak~Lermi, Max Molina~Ayala, Sharia Hernandez, Wei Lu, Khaja Khan, Alejandra Serrano, Idania Lubo, Leticia Hamana, Katarzyna Tomczak, Sean Barnes, et~al.
\newblock Comparison of imaging-based single-cell resolution spatial transcriptomics profiling platforms using formalin-fixed, paraffin-embedded tumor samples.
\newblock \emph{bioRxiv}, pages 2024--12, 2024.

\bibitem[Pan et~al.(2025)Pan, Salvatierra, Ercan, Kakarala, Lu, Shi, Lubo~Julio, Wistuba, Solis~Soto, and Yuan]{pan2025tmeseg}
Xiaoxi Pan, Maria~E Salvatierra, Caner Ercan, Lakshmi Kakarala, Wei Lu, Ou Shi, Idania~C Lubo~Julio, Ignacio~I Wistuba, Luisa~M Solis~Soto, and Yinyin Yuan.
\newblock Tmeseg: Connecting histopathology with spatial transcriptomics through tumor microenvironment segmentation for lung cancer.
\newblock \emph{Cancer Research}, 85\penalty0 (8\_Supplement\_1):\penalty0 2426--2426, 2025.

\bibitem[Ru et~al.(2023)Ru, Huang, Zhang, Aldape, and Jiang]{ru2023estimation}
Beibei Ru, Jinlin Huang, Yu Zhang, Kenneth Aldape, and Peng Jiang.
\newblock Estimation of cell lineages in tumors from spatial transcriptomics data.
\newblock \emph{Nature Communications}, 14\penalty0 (1):\penalty0 568, 2023.

\bibitem[Theodoris et~al.(2023)Theodoris, Xiao, Chopra, Chaffin, Al~Sayed, Hill, Mantineo, Brydon, Zeng, Liu, et~al.]{theodoris2023transfer}
Christina~V Theodoris, Ling Xiao, Anant Chopra, Mark~D Chaffin, Zeina~R Al~Sayed, Matthew~C Hill, Helene Mantineo, Elizabeth~M Brydon, Zexian Zeng, X~Shirley Liu, et~al.
\newblock Transfer learning enables predictions in network biology.
\newblock \emph{Nature}, 618\penalty0 (7965):\penalty0 616--624, 2023.

\bibitem[Wang et~al.(2025)Wang, Du, Liu, Ouyang, Chen, and Lin]{wang2025m2ost}
Hongyi Wang, Xiuju Du, Jing Liu, Shuyi Ouyang, Yen-Wei Chen, and Lanfen Lin.
\newblock M2ost: Many-to-one regression for predicting spatial transcriptomics from digital pathology images.
\newblock In \emph{Proceedings of the AAAI Conference on Artificial Intelligence}, pages 7709--7717, 2025.

\bibitem[Xie et~al.(2023)Xie, Pang, Chung, Perciani, MacParland, Wang, and Bader]{xie2023spatially}
Ronald Xie, Kuan Pang, Sai Chung, Catia Perciani, Sonya MacParland, Bo Wang, and Gary Bader.
\newblock Spatially resolved gene expression prediction from histology images via bi-modal contrastive learning.
\newblock \emph{Advances in Neural Information Processing Systems}, 36:\penalty0 70626--70637, 2023.

\bibitem[Xu et~al.(2024)Xu, Usuyama, Bagga, Zhang, Rao, Naumann, Wong, Gero, Gonz{\'a}lez, Gu, et~al.]{xu2024whole}
Hanwen Xu, Naoto Usuyama, Jaspreet Bagga, Sheng Zhang, Rajesh Rao, Tristan Naumann, Cliff Wong, Zelalem Gero, Javier Gonz{\'a}lez, Yu Gu, et~al.
\newblock A whole-slide foundation model for digital pathology from real-world data.
\newblock \emph{Nature}, 630\penalty0 (8015):\penalty0 181--188, 2024.

\bibitem[Zhang et~al.(2024)Zhang, Schroeder, Yan, Yang, Hu, Lee, Cho, Susztak, Xu, Feldman, et~al.]{zhang2024inferring}
Daiwei Zhang, Amelia Schroeder, Hanying Yan, Haochen Yang, Jian Hu, Michelle~YY Lee, Kyung~S Cho, Katalin Susztak, George~X Xu, Michael~D Feldman, et~al.
\newblock Inferring super-resolution tissue architecture by integrating spatial transcriptomics with histology.
\newblock \emph{Nature biotechnology}, 42\penalty0 (9):\penalty0 1372--1377, 2024.

\bibitem[Zhao et~al.(2024)Zhao, Liang, Huang, Huang, and Yu]{zhao2024hist2cell}
Weiqin Zhao, Zhuo Liang, Xianjie Huang, Yuanhua Huang, and Lequan Yu.
\newblock Hist2cell: Deciphering fine-grained cellular architectures from histology images.
\newblock \emph{bioRxiv}, pages 2024--02, 2024.

\end{thebibliography}
}

\end{document}